\documentclass[11pt,oneside,a4paper]{article}
\usepackage{ws-rv-van} 
\usepackage{authblk}
\usepackage{amsthm}
\usepackage{amssymb}
\usepackage{amsmath}
\usepackage{hyperref}
\usepackage{textcomp}
\usepackage{makecell}
\usepackage{xcolor}

\definecolor{darkgray}{gray}{0.25}

\usepackage{blindtext,tikz} 
\usetikzlibrary{calc}

\usepackage[draft]{todonotes}

\makeindex

\begin{document}

\title{Conversational Agents: Theory and Applications}
\author{Mattias Wahde}
\author{Marco Virgolin}
\affil{Department of Mechanics and Maritime Sciences\\
Chalmers University of Technology, 412 96 G\"oteborg, Sweden\\
\{mattias.wahde,marco.virgolin\}@chalmers.se}
\date{}

\maketitle

\tikz[overlay,remember picture]
{
    \node at ($(current page.south)+(0,+19)$) [rotate=0] {\scriptsize%
    \color{darkgray}
    \makecell{Preprint of a chapter published in Handbook of Computer Learning and Intelligence -- Volume 1\\
    \url{https://doi.org/10.1142/9789811246050_0012}\\
    \textcopyright 2022 World Scientific Publishing Company
    }};
}

\begin{abstract}
In this chapter, we provide a review of conversational agents (CAs), discussing 
chatbots, intended for casual conversation with a user, as well as task-oriented 
agents that generally engage in discussions intended to reach one or several specific goals, often (but not always) within a specific domain. We also consider the concept of embodied
conversational agents, briefly reviewing aspects such as character animation
and speech processing. The many different approaches for 
representing dialogue in CAs are discussed in some detail, 
along with methods for evaluating such agents, emphasizing the important
topics of accountability and interpretability. A brief historical overview
is given, followed by an extensive overview of various applications, 
especially in the fields of health and education. We end the chapter by
discussing benefits and potential risks regarding the societal impact
of current and future CA technology.
\end{abstract}


\tableofcontents

\section{Introduction}
\textit{Conversational agents} (CAs), also known as intelligent virtual agents (IVAs), are computer programs designed for natural conversation with human users, either involving informal chatting, in which case the system is usually referred to as a \textit{chatbot}, or with the aim of providing the user with relevant 
information related to a specific task (such as a flight reservation), in
which case the system is called a \textit{task-oriented agent}. 
Some CAs use only text-based input and output, whereas others involve more complex
input and output modalities (for instance, speech). There are also so-called embodied conversational agents (ECAs) that are typically equipped with an animated visual representation (face or body) on-screen. 
Arguably the most important part of CAs is the manner in which they represent and handle dialogue. 
Here, too, there is a wide array of possibilities, ranging from simple template-matching systems up to highly complex representations based on deep neural networks (DNNs).

Among the driving forces behind the current, rapid development of this field
is the need for sophisticated CAs in various applications, for
example in healthcare, education, and customer service; see also Section~\ref{sect:applications} below. 
In terms of research, the development is also, in part, driven by high-profile
competitions such as the Loebner prize and the Alexa prize, which are briefly
considered in Sections~\ref{sec:evaluation} and~\ref{sect:history}. 
Giving a full description of the rapidly developing field of CAs, with all of the facets introduced above, is a daunting task.
Here, we will attempt to cover both theory and applications, but we
will mostly focus on the representation and implementation of
dialogue capabilities (rather than, say, issues related to speech
recognition, embodiment, etc.), with emphasis on task-oriented
agents, particularly applications of such systems. 
We strive
to give a bird's eye view of the many different approaches that are,
or have been, used for developing the capabilities of CAs.
Furthermore we discuss the ethical implications of CA technology, especially
aspects related to interpretability and accountability.
The interested reader may also wish to consult other reviews on CAs and their applications\cite{LesterEtAl2004, MascheLe2017, LaranjoEtAl2018, DiederichEtAl2019,
BavarescoEtAl2020}.

\section{Taxonomy}
\label{sect:taxonomy}
The taxonomy used here primarily distinguishes between, on the one hand,
\textit{chatbots} and, on the other, \textit{task-oriented agents}. 
Chatbots are systems that (ideally) can maintain a casual dialogue with 
a user on a wide range of topics, but are generally neither equipped
not expected to provide precise information. 
For example, many (though not all) chatbots may give different answers if asked the same specific question several times (such as \lq\lq\textit{Where were you born?}\rq\rq).
Thus, a chatbot is primarily useful for conversation on everyday topics, such as 
restaurant preferences, movie reviews, sport discussions, and so on, where the actual content of the conversation is perhaps less relevant than the interaction itself: A 
well-designed chatbot can provide stimulating interactions for a human user, but would 
be lost if the user requires a detailed, specific answer to questions 
such as \lq\lq\textit{Is there any direct flight to London on Saturday morning?}\rq\rq.

Task-oriented agents, on the other hand, are ultimately intended to provide 
clear, relevant, and definitive answers to specific queries, a process that 
often involves a considerable amount of database queries (either offline or via the internet) and data processing that, with some generosity, can be referred to as \textit{cognitive processing}. 
This might be a good point to clear up some confusion surrounding
the taxonomy of CAs, where many authors refer to \textit{all}
such systems as chatbots. 
In our view, this is incorrect. As their name implies, chatbots do just that:~\textit{chat}, while task-oriented agents are normally used for more serious and complex tasks.

While these two categories form the basis for a classification of CAs, many other taxonomical aspects could be considered as well, for example the input and output modalities used (e.g.,~text, speech etc.), the applicability of the agent (general-purpose or domain-specific), whether or not the agent is able to self-learn, and so on\cite{DiederichEtAl2019}.
One such aspect concerns the representation 
used, where one frequently sees a division into rule-based systems 
and systems based on artificial intelligence (AI), by which is commonly 
meant systems based on deep 
neural networks (DNNs)\cite{LecunEtAl2015}.
This, too, is an unfortunate classification in our view; first of all,
the field of AI is much wider in scope than what the current focus on
DNNs might suggest: For example, rule-based systems, as understood in this context, have
played, and continue to play, a very important part in the field. Moreover,
even though many so-called rule-based agents are indeed based on 
\textit{handcrafted} rules, there is nothing preventing the use of 
machine learning methods (a subfield of AI), such as stochastic optimization methods\cite{Wahde2008}
or reinforcement learning\cite{Sutton2018}, in such systems. 
The name \textit{rule-based} is itself a bit unfortunate, since a \textit{rule} is a very generic concept that could, at least in principle, even be applied to the individual components of DNNs.
Based on these considerations, we will here suggest an alternative dichotomy, which will be used alongside the other classification given above. This alternative classification divides CAs (whether they are chatbots or task-oriented agents) into the two categories \textit{interpretable} systems and \textit{black box} systems. 
We hasten to add that \textit{interpretable} does not imply a lack of complexity: In principle, the behavior of an interpretable CA could be every bit as sophisticated as that of a black box (DNN) CA. 
The difference is that the former are based on 
transparent structures consisting of what one might call \textit{interpretable primitives}, i.e., human-understandable components such as IF-THEN-ELSE-rules, sorting functions, or AIML-based statements (see Section~\ref{subsubsect:patternbased}), which directly carry out high-level operations without the need to rely on the concerted action of a huge number of parameters as in a DNN.

As discussed in Sections~\ref{sec:evaluationSocietalImplications} and~\ref{sect:future} below, this is a distinction that, at least for task-oriented agents, is likely to become increasingly important in the near future: CAs are starting to be applied in high-stakes decisions, where interpretability is a crucial aspect\cite{Rudin2019}, not least in view of recently proposed legislation (both in the EU and the USA) related to a user's right to an explanation or
motivation, for example in cases where artificial systems are involved
in bank credit decisions or clinical decision-making\cite{GoodmanFlaxman2017}. 
In such cases,
the artificial system must be able to explain its reasoning or at least
carry out its processing in a manner that can be understood by a human operator.

\section{Input and output modalities}\label{sec:input-output modalities}
Textual processing, which is the focus of the sections below, is arguably the core functionality of CAs.
However, natural human-to-human conversation also involves aspects such as speech, 
facial expressions, gaze, body posture, and gestures, all of which convey subtle but
important nuances of interaction beyond the mere exchange of
factual information. In other words, interactions between
humans are \textit{multimodal}.

\begin{figure}
\centering
    \includegraphics[width=\linewidth]{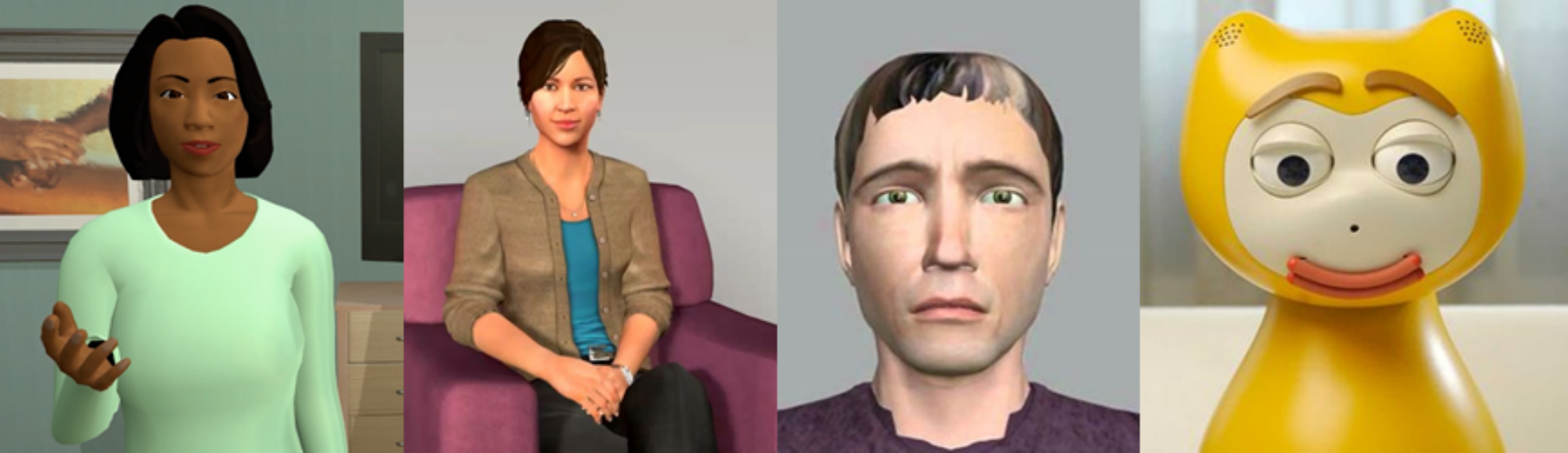}
\caption{Three examples of ECAs, and one social robot (right). From left to right: Gabby~\cite{GardinerEtAl2017,JackEtAl2020}, SimSensei~\cite{deVaultEtAl2014}, Obadiah\cite{OchsEtAl2017}, and iCAT \cite{vanBreemen2005} (Royal Philips / Philips Company Archives). All images are reproduced with kind permission from the respective copyright holders.}
\label{fig:ecas}
\end{figure}
Aspects of multimodality are considered in the field of \textit{embodied conversational agents} (ECAs, also known as \textit{interface agents})
that involve an interface for natural interaction with the user.
This interface is often in the form of an animated face (or body) shown 
on-screen, but can also be a physical realization in the form of a
\textit{social robot}\cite{LeiteEtAl2013,BelpaemeEtAl2018,AnzaloneEtAl2015}.
Figure~\ref{fig:ecas} gives some examples; the three left
panels show agents with a virtual (on-screen) interface, whereas the rightmost panel shows the \textit{social robot} iCAT.
Here, we will focus on virtual interfaces rather than physical implementations such as robots.

\subsection{Non-verbal interaction}
Humans tend to anthropomorphize systems that have a life-like shape, such as an animated face on-screen\cite{EpleyEtAl2008}. 
Thus, a user interacting with an ECA may experience better rapport with the agent than would be the case for an interaction with an agent lacking a visual representation, and the embodiment also increases (in many, though not all, cases) the user's propensity to trust the agent, as found by Loveys~\textit{et al.}\cite{LoveysEtAl2020}. 
Some studies show that certain aspects of an ECA, such as presenting a smiling face or showing a
sense of humor, improves the experience of interacting with the
agent\cite{CreedBeale2012,KulmsEtAl2014}, even though the \textit{user's}
personality is also a relevant factor when assessing the quality
of interaction. 
For example, a recent study by ter Stal~\textit{et al.} on ECAs in the context of eHealth showed that users tend to prefer ECAs that are similar to 
themselves regarding age and sex (gender)\cite{terStalEtAl2020}.
Other studies show similar results regarding personality traits
such as extroversion and introversion\cite{vonDerPuttenEtAl2010,CerekovicEtAl2014}.
Additional relevant cues for human-agent interaction include 
gaze\cite{RuhlandEtAl2015,PejsaEtAl2015}, body posture\cite{MarschnerEtAl2015}
and gestures\cite{Pelachaud2009,SadeghipourKopp2011}, as well as the interaction between these expressive modalities\cite{MarschnerEtAl2015}.

Another point to note is that \textit{human} likeness is not a prerequisite for establishing rapport between a human user and an ECA\cite{BergmannEtAl2012}. 
In fact, to some degree, one may argue that the opposite holds: Mori's \textit{uncanny valley} hypothesis\cite{MoriEtAl2012} states essentially that artificial systems that attempt to mimic human features in great detail, without fully succeeding in doing so, tend to be perceived as eerie and repulsive. 
This hypothesis is supported by numerous studies in (humanoid) robotics, but also in the context of virtual agents\cite{TinwellEtAl2011}.
What matters from the point of view of user-agent rapport seems instead to be that the appearance of an ECA should match the requirements of its task\cite{KeelingEtAl2004}, 
and that the ECA should make use of several interaction modalities\cite{BergmannEtAl2012}.

Conveying emotions is an important purpose of facial expressions. 
Human emotions are generally described in terms of a set of basic emotions. 
Ekman\cite{Ekman1999} defined six basic emotions, namely anger, disgust, fear, happiness, sadness, and surprise, a set that has later been expanded by others, to cover additional basic emotions as well as linear combinations thereof\cite{Whissell1989}.
Emotions can be mapped to facial expressions via the facial action coding system (FACS)\cite{Hjortsjo1969, EkmanFriesen1978} that, in turn, relies on so-called action units (AUs) which are directly related to movements of facial muscles (as well as movements of the head and the eyes). 
For example, happiness can be modelled as the combination of AU6 (\textit{cheek raiser}) and AU12 (\textit{lip corner puller}). 
Emotion recognition can also be carried out using black box systems like DNNs\cite{LiDeng2020}, trained \emph{end-to-end}: These systems learn to predict the emotion from the image of a face, without the need to manually specify how intermediate steps should be carried out\cite{Ko2018}. State-of-the art facial emotion recognition systems typically achieve 80 to 95\% accuracy over 
benchmark data sets\cite{MehtaEtAl2018}.

Emotions are also conveyed in body language\cite{Coulson2004}, albeit with greater cultural variation than for the case of facial expressions. ECAs equipped with cameras
(especially depth cameras\cite{dOrazio2016EtAl}) can be made to 
recognize body postures, gestures, and facial expressions (conveying emotional states),
and to react accordingly. Gesture recognition, especially in the case of hand gestures, helps to achieve a more natural interaction between the user and the ECA\cite{RautarayAgrawal2015}. Furthermore, the combination of audio and visual clues can be exploited to recognize emotions by multimodal systems\cite{KahouEtAl2016,FayekEtAl2017}. 

\subsection{Character animation}
Methods for character animation can largely be divided into three categories.
In \textit{procedural animation} the poses of an animated body or face are parameterized, using underlying skeletal and muscular models (rigs)\cite{GilliesSpanlang2010} coupled with inverse kinematics. 
In procedural \textit{facial} animation\cite{SerraEtAl2018} every unit of sound (phoneme) is mapped to a corresponding facial expression (a \textit{viseme}).
Animations are then built in real time, either by concatenation or blending, a process that allows full control over the animated system, but is also computationally demanding and may struggle to generated life-like animations.
In \textit{data-driven animation}, the movement sequences are built by stitching together poses (or facial expressions) from a large database.
Many different techniques have been defined, ranging from simple linear interpolations to methods that involve DNNs\cite{ZhangEtAl2018, LeeEtAl2018}.
Finally, \textit{motion capture-based} (or \textit{performance-based}) methods\cite{WeiseEtAl2011} map the body or facial movements of a human performer onto a virtual, animated character. 
This technique typically generates very realistic animations, but is also laborious and costly. 
Moreover, the range of movements of the virtual character is also limited by the range of expressions generated by the human performer\cite{EdwardsEtAl2016}.
Several frameworks have been developed for animating ECAs. 
An example is the SmartBody approach by Thiebaux~\textit{et al.}\cite{ThiebauxEtAl2008}  that makes use of the behavior markup language (BML)\cite{KoppEtAl2006} to generate animation sequences that synchronize facial motions and speech.

\subsection{Speech processing}
Speech processing is another important aspect of an ECA or, more generally, of a \textit{spoken
dialogue system} (SDS)\footnote{An SDS is essentially a CA that processes speech (rather than text) as its input and output.}.
CAs handle incoming information in a step involving \textit{natural language
understanding} (NLU; see also Figure~\ref{fig:pipelinemodel} below). 
In cases where speech (rather
than just text) is used as an input modality, NLU is preceded by \textit{automatic speech recognition} (ASR). The ASR step may provide a more natural type of interaction from the user's point of view, but it also increases the complexity of the implementation considerably, partly  because of the inherent difficulty in reliably recognizing speech, and partly because spoken language tends to contain aspects that are not generally present
in textual input (e.g.,~repetition of certain words, the use of interjections
such as \lq\lq\textit{uh}\rq\rq, \lq\lq\textit{er}\rq\rq and so on, as well as noise on various levels).

The performance of ASR can be measured in different ways: A common measure is the 
word error rate (WER), defined as
\begin{equation}
{\rm{WER}} = \frac{S+I+D}{N},
\end{equation}
where $N$ is the number of words in the reference sentence (the ground truth), and $S$, $I$, and $D$ are, 
respectively, the number of (word) substitutions, insertions, and deletions required to go from the 
hypothesis generated by the ASR system to the reference sentence. 
It should be noted, however, that the WER does not
always represent the true ASR performance accurately, as it does not capture the
semantic content of an input sentence. For example, in some cases, several words
can be removed from a sentence without changing its meaning whereas in other
cases, a difference of a single word, e.g.,~omitting the word \textit{not}, may completely change the meaning.
Still, the WER is useful in \textit{relative} performance assessments of different ASR methods.

With the advent of deep learning, the accuracy of ASR has increased considerably. 
State-of-the-art methods for ASR where DNNs are used both for phoneme recognition and for word decoding, generally outperform\cite{KepuskaBohouta2017} earlier systems that
were typically based on hidden Markov models (HMMs).
For DNNs, WERs of 5\% or less (down from around 10-15\% in the early 2010s) have been reported\cite{ChiuEtAl2018,ParkEtAl2019}, 
implying performance on a par with, or even exceeding, human performance.
However, it should be noted that these low WERs generally pertain to low-noise
speech in benchmark data sets, such as the LibriSpeech data set which is 
based on audio books\cite{PanayotovEtAl2015}. In practical applications,
with noisy speech signals, different speaker accents, and so on, ASR performance 
can be considerably worse\cite{AmodeiEtAl2016}.

Modern text-to-speech (TTS) synthesis, which also relies on deep learning to a
great degree, generally reaches a high level
of naturalness\cite{vanDenOordEtAl2018, BinkowskiEtAl2019}, as measured (subjectively) 
using the five-step mean opinion score (MOS) scale\cite{StreijlEtAl2016}.
The requirements on speech output varies between different systems.
For many ECAs, speech output must not only transfer information in a clear manner but
also convey emotional states that, for example, match the facial expression of the ECA.
It has been demonstrated that the pitch of an ECA voice influences
the agent's perceived level of trustworthiness\cite{ElkinsDerrick2013}. Similar
results have been shown for ECAs that speak with a happy voice\cite{TorreEtAl2020}.

Finally, with the concept of \textit{augmented reality}, or the still more ambitious concept of \textit{mixed reality},
even more sophisticated levels of embodiment are made possible\cite{AnabukiEtAl2000}, where 
the embodied agent can be superposed onto the real world, for an immersive user experience.
This approach has recently gained further traction\cite{WangEtAl2019b} with the advent of
wearable augmented reality devices such as Microsoft's Hololens and Google Glass.

\section{Dialogue representation}
A defining characteristic of any CA is the manner in which it handles
dialogue. In this section, dialogue representation is reviewed both for
chatbots and task-oriented agents, taking into consideration the fact
that chatbots handle dialogue in a different and more direct manner
than task-oriented agents. 

\subsection{Chatbots}
Chatbots can be categorized into three main groups by the manner in which they generate their responses. Referring also to the alternative dichotomy
introduced in Section~\ref{sect:taxonomy}, the first two types,
namely \textit{pattern-based chatbots} and \textit{information-retrieval chatbots}, would fall into the category of interpretable systems, whereas the third
type, \textit{generative chatbots}, are either of the interpretable
kind or the black box variety depending on the implementation used; those that make strong use of DNNs would cleanly fall into the black box category.

\subsubsection{Pattern-based chatbots}
\label{subsubsect:patternbased} 
The very first CA, namely Weizenbaum's ELIZA\cite{Weizenbaum1966} belongs to
a category that can be referred to as 
\textit{pattern-based} chatbots\footnote{This type of chatbot is also commonly
called \textit{rule-based}. However, one might raise the same objections 
to that generic term as expressed in Section~\ref{sect:taxonomy}
above, and therefore the more descriptive term \textit{pattern-based} will
be used instead.}. This chatbot, released in 1966, was meant to emulate
a psychotherapist who attempts to gently guide a patient along a 
path of introspective discussions, often by transforming and reflecting
statements made by the user. ELIZA operates by matching
the user's input to a set of patterns, and then applying 
cleverly designed rules to formulate its response. Thus, as a simple 
example, if a user says \lq\lq\textit{I'm depressed}\rq\rq, ELIZA could use the template
\texttt{rule (I'm 1)} $\rightarrow$ \texttt{(I'm sorry to hear that you are 1)}, where
the \texttt{1} indicates a single word, to return \lq\lq\textit{I'm sorry to hear 
that you are depressed}\rq\rq. 
In other cases, where no direct match can be
found, ELIZA typically just urges the user to continue. For
example, the input \lq\lq\textit{I could not go to the party}\rq\rq can be
followed by \lq\lq\textit{That is interesting. Please continue}\rq\rq 
or \lq\lq\textit{Tell me more}\rq\rq. ELIZA also ranks patterns, such that,
when several options match the user's input, it will produce
a response based on the highest-ranked (least generic) pattern.
Furthermore, it features a rudimentary short-term memory, allowing it
to return to earlier parts of a conversation, to some degree.

More modern pattern-based chatbots, some of which are discussed
in Section~\ref{sect:history} below, are often based on the
Artificial Intelligence Markup Language (AIML)\cite{Wallace2003}, an XML-like
language developed specifically for defining template-matching
rules for use in chatbots. Here, each pattern (user input) is
associated with an output (referred to as a \textit{template}).
A simple example of an AIML specification is
\begin{verbatim}
<category>
 <pattern> I LIKE *</pattern>
 <template>I like <star/> as well</template>
<\category>
\end{verbatim}
With this specification, if a user enters \lq\lq\textit{I like tennis}\rq\rq,
the chatbot would respond by saying \lq\lq\textit{I like tennis as well}\rq\rq.
AIML has a wide range of additional features, for example those
that allows it to store, and later retrieve, variables (e.g.,~the user's name).
A chatbot based on AIML can also select random responses from a set of candidates, 
so as to make the dialogue more lifelike.
Moreover, AIML has a procedure for making
redirections from one pattern to another, thus greatly
simplifying the development of chatbots. For example,
with the specification
\begin{verbatim}
<category>
 <pattern>WHAT IS YOUR NAME</pattern>
 <template>My name is Alice</template>
<\category>
<category>
 <pattern>WHAT ARE YOU CALLED</pattern>
 <template>
  <srai>WHAT IS YOUR NAME</srai>
 </template>
<\category>
\end{verbatim}
the agent would respond \lq\lq\textit{My name is Alice}\rq\rq if asked
\lq\lq\textit{What is your name}\rq\rq or \lq\lq\textit{What are you called}\rq\rq.
As can be seen in the specification, the second pattern
redirects to the first, using the so-called \textit{symbolic reduction
in artificial intelligence} (\texttt{srai}) tag.

\subsubsection{Information-retrieval chatbots}
Chatbots in this category generate their responses by selecting a suitable sentence from a (very) large dialogue corpus, i.e.~a database of stored conversations.
Simplifying somewhat, the basic approach is as follows: The agent (1) receives a user input $\mathcal{U}$; (2) finds the most similar sentence $\mathcal{S}$ (see below) in the associated dialogue corpus; and (3) responds with the sentence $\mathcal{R}$ that, in the dialogue corpus, was given in response to the sentence $\mathcal{S}$.

Similarity between sentences can be defined in different ways. Typically, sentences
are encoded in the form of numerical vectors, called \emph{sentence embeddings}, such that numerical measures of similarity can be applied.
A common approach is to use TF-IDF\cite{SaltonBuckley1988} (short for \textit{term frequency - inverse document frequency}),
wherein each sentence is represented by a vector whose length equals the number of words\footnote{Usually after stemming and lemmatization, i.e.~converting words to their basic form so that, for example, \textit{cats} is represented as \textit{cat}, \textit{better} as \textit{good}, and so on.} ($N$) available in the dictionary used. The TF vector is simply the frequency of occurrence of each word, normalized by the number of words in the sentence. Taken alone, this measure tends to give too much emphasis to common words such as \lq\lq\textit{the}\rq\rq, \lq\lq\textit{is}\rq\rq, \lq\lq\textit{we}\rq\rq, which do not really say very much about the content of the sentence. 
The IDF for a given word $w$ is typically defined as
\begin{equation}
\text{IDF} = \log \frac{M}{M_w},
\end{equation}
where $M$ is the total number of sentences (or, more generally, documents) in the corpus and 
$M_w$ is the number of sentences that contain $w$. Thus, IDF will give high values to words that are infrequent in the dialogue corpus and therefore likely to carry more relevant information about the contents of a sentence than more frequent words. The IDF values can be placed in a vector of length $N$ that, for a fixed dialogue corpus, can be computed once and for all. Next, given the input sentence $\mathcal{U}$, a vector $q(\mathcal{U})$ can be formed by first generating the corresponding TF vector and then performing a Hadamard (element-wise) product with the IDF vector. 
The vector $q(\mathcal{U})$ can then be compared with the corresponding vector $q(\sigma_i)$ for all sentences $\sigma_i$ in the corpus,
using the vector (dot) product, and then extracting the most similar sentence $\mathcal{S}$ by computing the maximum cosine similarity as
\begin{equation}
\mathcal{S} = \text{argmax}_i\left(\frac{q(\mathcal{U})\cdot q(\sigma_i)}{\Vert q(\mathcal{U}) \Vert 
\Vert q(\sigma_i)\Vert}\right).
\label{eq:tfidf}
\end{equation}
The chatbot's response is then given as the response $\mathcal{R}$ to $\mathcal{S}$ in the dialogue corpus. An alternative approach is to directly retrieve the response that best matches the user's input~\cite{Ritter2011}.

Some limitations of the TF-IDF approach are that it does not take into account (1) the order in which the words appear in a sentence (which of course can have a large effect on the semantic content of the sentence) and (2) the fact that many words have synonyms. Consider, for instance, the two words \textit{mystery} and \textit{conundrum}. A sentence containing one of those words would have a term frequency of 1 for that word, and 0 for the other, so that the two words together give a zero contribution to the cosine similarity in Equation~(\ref{eq:tfidf}). Such problems can be addressed using \textit{word embeddings}, which are discussed below.
Another limitation is that TF-IDF, in its basic form, does not consider the \textit{context} of the conversation. Context handling can be included by, for example, considering earlier sentences in the conversation\cite{ZhaoEtAl2019}.

\begin{figure}
    \centering
    \includegraphics[width=0.7\linewidth]{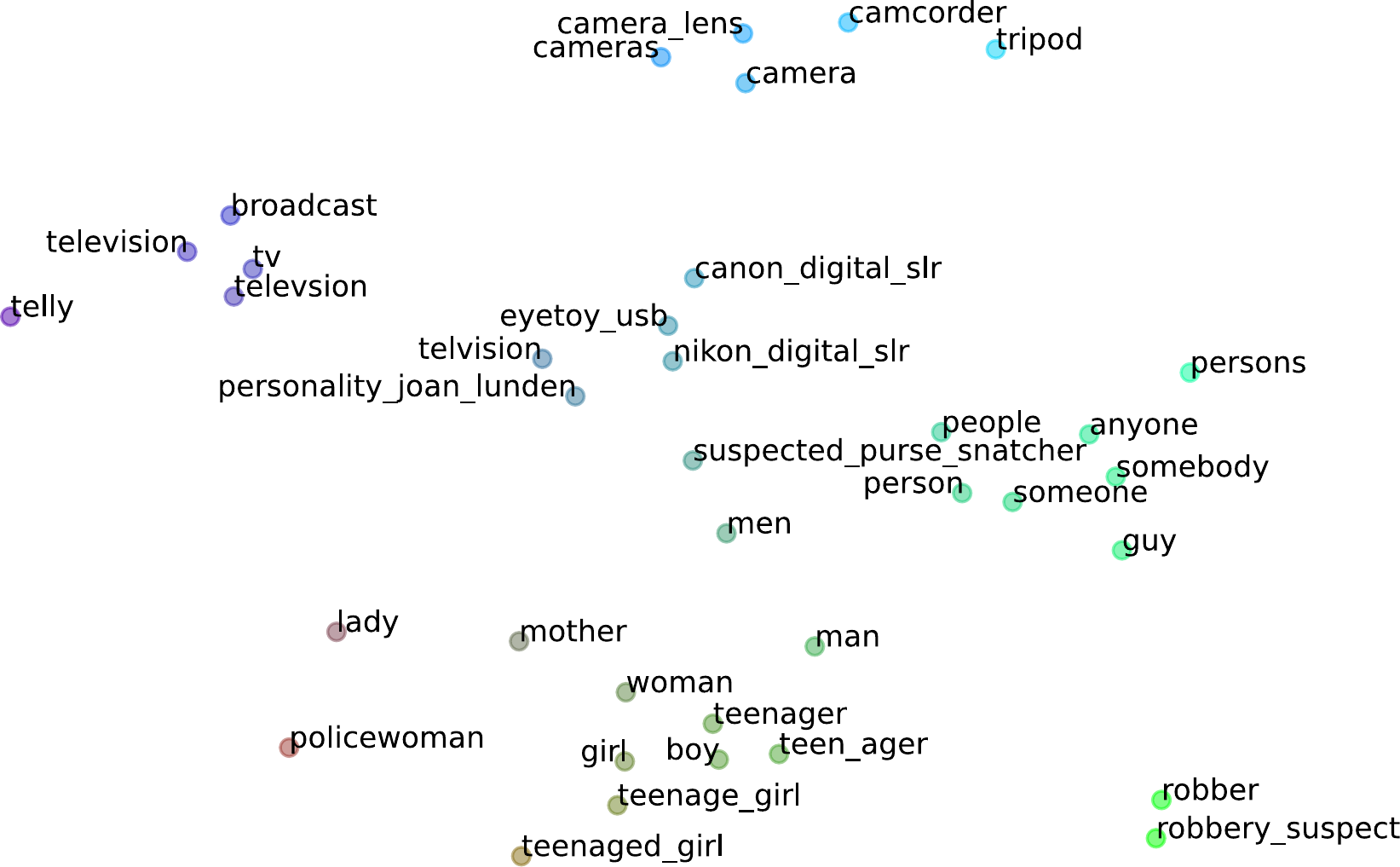}
    \caption{Projection obtained with t-SNE\cite{MaatenHinton2008} of almost forty $300$-dimensional word embeddings from word2vec\cite{MikolovChenEtAl2013}. 
    The projection mostly preserves embedding proximity, as similar words are clustered together. 
    A notable exception is \texttt{suspected\_purse\_snatcher}, that is in the center of the plot and far from \texttt{robber} and \texttt{robbery\_suspect}, in the bottom right. 
    Note the presence of misspelled words from natural text, such as \texttt{telvision} and \texttt{televsion}.
    }
    \label{fig:wordembeddings}
\end{figure}

To overcome the limitations of TF-IDF, Yan \textit{et al.}~proposed an information-retrieval chatbot where a DNN is used to consider context from previous exchanges, and rank a list of responses retrieved by TF-IDF from least to most plausible\cite{YanEtAl2016}.
Other works on information-retrieval chatbots via DNNs abandon the use of TF-IDF altogether, and focus on how to improve the modeling of contextual information processing\cite{LoweEtAl2015,ZhouEtAl2016,WuEtAl2019}.
An aspect that is common to different types of DNN-based chatbots (of any kind) and incidentally also task-oriented agents, is the use of \emph{word embeddings}.
Similar to sentence embeddings, an embedding $e_w$ of a word $w$ is a vectorial representation of $w$, i.e., $e_w \in \mathbb{R}^d$. 
DNNs rely on word embeddings in order to represent and process words, and can actually learn the embeddings during their training process\cite{BengioEtAl2003,Schwenk2007,MikolovEtAl2013,DevlinEtAl2019}.
Typically, this is done by incorporating an $N \times d$ real-valued matrix $\Omega$ within the structure of the DNN as a first layer, where $N$ is the number of possible words in the vocabulary (or, for long words, word-chunks) and $d$ is the (predetermined) size of each word embedding.
$\Omega$ is initialized at random. 
Whenever the network takes some words $w_i, w_j, w_k, \dots$ as input, only the corresponding rows of $\Omega$, i.e., the embeddings $e_{w_i}, e_{w_j}, e_{w_k}, \dots$, become active and pass information forward to the next layers.
During the training process, the parameters of the DNN, which include the values of $\Omega$ and thus of each $e_{w_i}$, are updated. 
Ultimately, each word embedding will capture a particular meaning.
For example, Mikolov \textit{et al.}\cite{MikolovEtAl2013} famously showcased that arithmetic operations upon DNN-learned word embeddings can be meaningful: 
\begin{equation}
    e_\text{King} - e_\text{Man} + e_\text{Woman} \approx e_\text{Queen}.
\end{equation}
As an additional example, Figure~\ref{fig:wordembeddings} shows a word cloud obtained by applying dimensionality reduction to a few dozen $300$-dimensional word embeddings. 
As can be seen, embeddings of words with similar syntax or semantics are clustered together.

We remark that other ways than training DNNs exist to build word embeddings, resulting in word embeddings that carry information of other nature. Some of these methods are focused, e.g., on statistical occurrences of words within document classes\cite{DumaisEtAl1988,LandauerEtAl1998}, or on word-word co-occurrences\cite{LundBurgess1996,Collobert2014}.
Moreover, word embeddings can be used as numerical word representations in other systems, e.g., in evolutionary algorithms, to obtain human-interpretable mathematical expressions representing word manipulations\cite{ManzoniEtAl2020}.

\subsubsection{Generative chatbots}\label{sec:generative-chatbots}
While chatbots in the two previous categories rely on existing utterances, either in the form of pre-specified patterns or retrieved from a dialogue corpus, generative chatbots instead \textit{generate} their responses using statistical models, called \emph{generative models} or, specifically when one intends to model probability distributions over language (e.g., in the form of what words are likely to appear after or in between some other words), \emph{language models}. 
Currently, this field is dominated by black box systems realized by DNNs trained on large amounts of data.

\begin{figure}
    \centering
    \includegraphics[width=\linewidth]{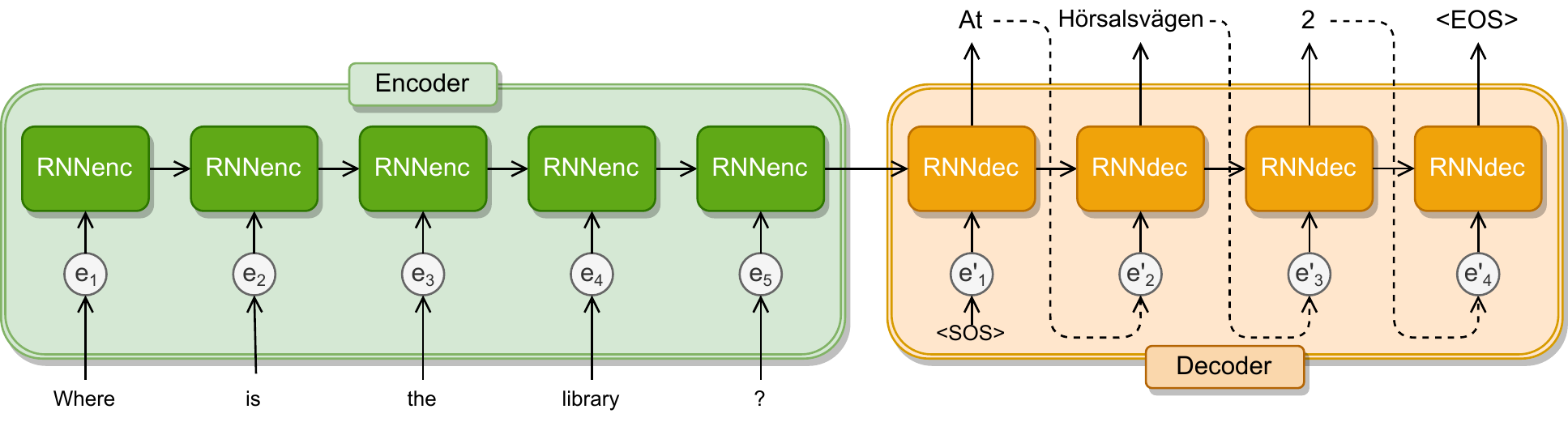}
    \caption{Simplified representation of seq2seq, unfolded over time. Each token of the input sentence \lq\lq\emph{Where is the library?}\rq\rq is transformed into an embedding and fed to the recurrent neural network called the \emph{encoder} (RNN\textsubscript{enc}). This information is passed to the recurrent neural network responsible for providing the output sentence, called the \emph{decoder} (RNN\textsubscript{dec}). Note that each output token is routed back in the decoder to predict the next output token. The special tags \texttt{\textlangle SOS\textrangle} and \texttt{\textlangle EOS\textrangle} identify the start and end of the sentence, respectively.}
    \label{fig:seq2seq}
\end{figure}

An important neural network model that is used at the heart of several generative chatbots is the \emph{sequence to sequence} model (seq2seq)\cite{SutskeverEtAl2014}. 
Figure~\ref{fig:seq2seq} shows a simplified representation of seq2seq. 
Given a sequence of general tokens (in this case, words), seq2seq returns a new sequence of tokens, by typically making use of recurrent neural networks. 
Generally speaking, these networks operate by taking as input, one by one, the tokens that constitute a given sequence; processing each token; and producing an output that, crucially, depends on both the token processing and the outputs produced so far.
Depending on the task, the training process of these systems can be based on the feedback of interacting users (a feedback that can be as simple as just indicating whether answers are \emph{good} or \emph{bad}) or on reference output text, such as known translations, answers to questions, or phrases where some words are masked and need to be guessed.

One of the first uses of seq2seq to realize (part of the conversational capability of) chatbots can be attributed to Vinyals and Quoc\cite{VinyalsQuoc2015}, who showed that even though seq2seq was originally proposed for machine translation, it could be trained to converse when provided with conversation corpora (e.g., movie subtitles\cite{Tiedemann2009}).
Today, seq2seq is a core component in several state-of-the-art generative chatbots:  MILABOT\cite{SerbanEtAl2017} from the Montreal Institute for Learning Algorithms, Google's Meena\cite{AdiwardanaEtAl2020}, and Microsoft's XiaoIce\footnote{Note that XiaoIce is a large system that includes many other functionalities than chatting.}\cite{ZhouEtAl2020}.

Another recent and popular component of DNNs useful for generating language is the \emph{transformer}\cite{VaswaniEtAl2017,WolfEtAl2020}. 
Like recurrent neural networks, the transformer is designed to process sequential data, but it is different in that it does not need to parse the sequence of tokens in a particular order.
Rather, the transformer leverages, at multiple stages, an information retrieval-inspired 
mechanism known as \emph{self-attention}.
This mechanism makes it possible to consider context from any other position in the sentence\footnote{The concept of (self-)attention is not a transformer-exclusive mechanism, in fact, it was firstly proposed to improve recurrent neural networks\cite{BahdanauEtAl2015,LuongEtAl2015}.}.
Broadly speaking, self-attention is capable of scoring, for each token (or better, for each embedding), how much contextual focus should be put on the other tokens (embeddings), no matter their relative position. 
For example, in the phrase \lq\lq\emph{Maria left her phone at home}\rq\rq, the token \lq\lq\emph{her}\rq\rq will have a large attention score for \lq\lq\emph{Maria}\rq\rq, while \lq\lq\emph{home}\rq\rq will have large attention scores for \lq\lq\emph{left} and \lq\lq\emph{at}\rq\rq.
In practice, self-attention is realized by matrix operations, and the weights of the matrices are optimized during the training process of the DNN, just like any other component.

Transformers owe much of their popularity in the context of natural language processing to BERT (Bidirectional Encoder Representations from Transformers), proposed by Devlin \textit{et al.}\cite{DevlinEtAl2019}.
At the time of its introduction, BERT outperformed other contenders on popular natural language processing benchmark tasks of different nature\cite{WangEtAl2018}, at times by a considerable margin. 
Before tuning the DNN that constitutes BERT to the task at hand (e.g., language translation, question answering, or sentiment analysis), pre-training over large text corpora is carried out, to make the system learn what words are likely to appear next to others. 
Specifically, this is done by (1) collecting large corpora of text; (2) automatically masking out some tokens (here, words); and (3) optimizing the model to infer back the masked tokens. 
When trained enough, the self-attention mechanism is capable to gather information on the most important context for the missing tokens.
For example, when the blanks need to be filled in \lq\lq\emph{Maria \underline{\hspace{.3cm}} \underline{\hspace{.3cm}} phone at home}\rq\rq, a well-tuned self-attention will point to the preceding token \lq\lq\emph{Maria}\rq\rq to help infer \lq\lq\emph{her}\rq\rq, and point to \lq\lq\emph{at}\rq\rq and \lq\lq\emph{home}\rq\rq to infer \lq\lq\emph{left}\rq\rq.
In particular when trained to guess the token at the end of the sequence, transformer-based DNNs are proficient at \textit{generating} language. At inference time, they can take the user's utterance as the initial input.

Today's state-of-the-art DNNs for language modeling use a mix of (advanced) recurrent neural networks and transformers. 
Notable examples include ALBERT\cite{LanEtAl2019}, XLNet\cite{YangEtAl2019}, and the models by OpenAI, GPT2 and GPT3\cite{RadfordEtAl2019,BrownEtAl2020}.
The latter in particular contains hundreds of billions of parameters, was trained on hundreds of billions of words, and is capable to chat quite well even in so-called \emph{few}- or \emph{zero-shot} learning settings, i.e., respectively when a very limited number of examples, or no examples at all, are used to tune the model to chatting.
For the readers interested in delving into the details of DNNs for language modeling, we refer to the recent surveys by Young \textit{et al.}\cite{YoungEtAl2018} and Otter \textit{et al.}\cite{OtterEtAl2020}.

\begin{figure}[ht]
    \centering
    \includegraphics[width=\linewidth]{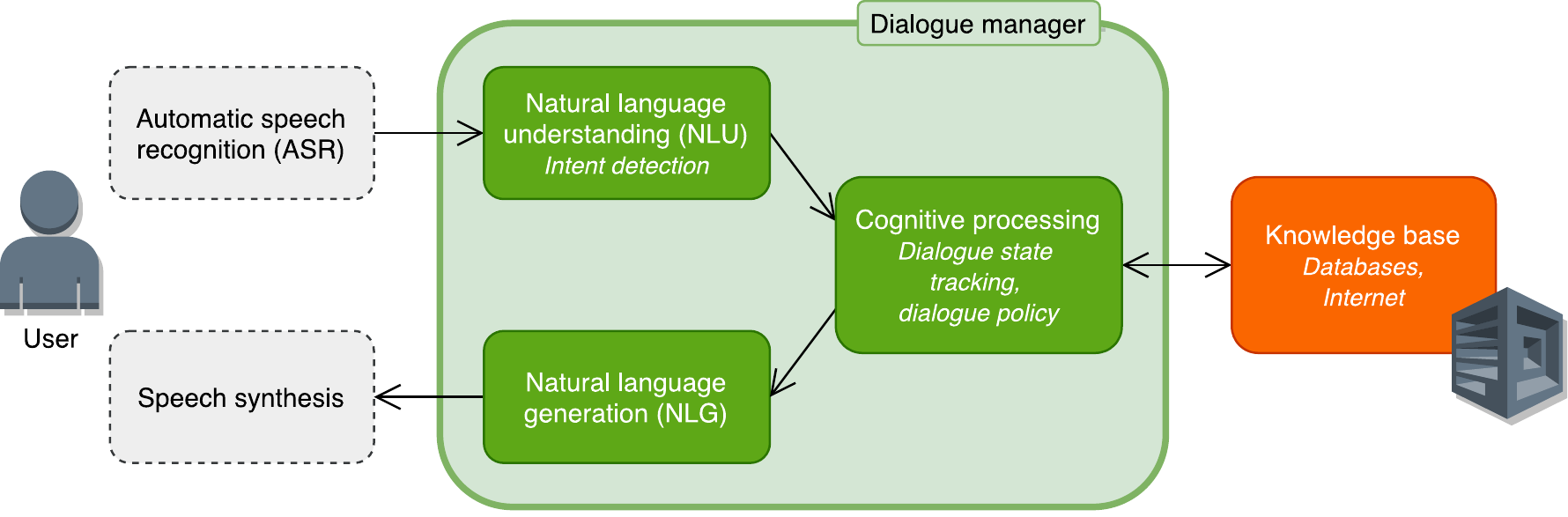}
    \caption{The pipeline model for CAs. The first and last steps (ASR and
    speech synthesis; see also Section~\ref{sec:input-output modalities} above) 
    can to some degree be considered as external from the core components, shown in green. It should also be noted that there are different versions of the pipeline model, with slightly different names for the various components, but nevertheless essentially following the structure shown here.}
    \label{fig:pipelinemodel}
\end{figure}

\subsection{Task-oriented agents}
\label{subsect:taskoriented}
As mentioned in Section~\ref{sect:taxonomy} above, the role
of task-oriented agents is to engage in fact-based
conversations on specific topics, and they must thus
be able to give precise, meaningful, and consistent answers to the user's 
questions. Here, we will review such agents, following the alternative
dichotomy from Section~\ref{sect:taxonomy}, where agents
are classified as either interpretable systems or black box
systems. The use of this dichotomy is particularly important here since task-oriented agents are responsible for reaching concrete goals and may be deployed in sensitive settings (for example, medical or financial applications; see also Section~\ref{sect:applications} below) where transparency and accountability are essential.

Figure~\ref{fig:pipelinemodel} shows the so-called \textit{pipeline
model}, in which the various stages of processing are shown as
separate boxes. 
In this model, the first and last steps, namely
ASR and speech synthesis, can perhaps be considered
to be peripheral, as many agents operate in a completely 
text-based fashion. Turning to the core elements, shown
in green in Figure~\ref{fig:pipelinemodel}, one can view
the processing as taking place in three stages: First,
in a step involving \textit{natural language understanding} (NLU)
the CA must identify precisely \textit{what} the user
said, or rather what the user \textit{meant}, hence
the notion of \textit{intent detection}. Next, in
a step that one may refer to as \textit{cognitive processing}\cite{Wahde2019},
the agent executes its \textit{dialogue policy}, i.e. undertakes 
the actions needed to construct the knowledge required for the response,
making use of a \textit{knowledge base} in the form of an
offline database, or via the internet or cloud-based services.
A key factor here is to keep track of context, also known as \textit{dialogue state}.
Finally, the CA formulates the answer as a sentence, in
the process of \textit{natural language generation} (NLG).

While the pipeline model offers a clear, schematic description of a
task-oriented agent, it should be noted that not all such agents
fall neatly into this model. 
Moreover, in the currently popular black box systems, the distinction between the stages tends to become blurred, as research efforts push towards requiring less and less human specification of how the processing should take place; see also Subsection~\ref{sect:blackbox} below.

\subsubsection{Interpretable dialogue systems}
\label{sect:interpretable}
The simplest examples of interpretable, task-oriented agents are
so-called \textit{finite-state} systems\cite{JurafskyMartin2009}, where the dialogue follows a rigid, predefined structure, such that 
the user is typically required to provide information in a given sequence. 
For example, if the CA is dealing with a hotel reservation, it might first ask the user \lq\lq\textit{Which city are you going to?}\rq\rq, then \lq\lq\textit{Which day will you arrive?}\rq\rq and so on, requiring an answer to each question in sequence. 
In such a system the dialogue initiative rests entirely with the CA, and the user has no alternative but to answer the questions in the order specified by the agent. 
In addition, finite-state systems can be complemented with the capability of handling \textit{universal commands} (such as \lq\lq\textit{Start over}\rq\rq) which the user might give at any point in the conversation. 

Even with such features included, however, CAs based on the finite-state approach are generally too rigid for any but the most basic kinds of dialogues; for example, in the case of a natural conversation regarding a reservation (flight, hotel, restaurant, etc.), the user may provide the required information in different order, sometimes even conveying several parts of the request in a single sentence (e.g., \lq\lq\textit{I want to book a flight to Paris on the $3^{\rm{rd}}$ of December}\rq\rq).
Such aspects of dialogue are handled, to a certain degree, in \textit{frame-based} (or \textit{slot-filling}) systems, such as GUS\cite{BobrowEtAl1977}, where the dialogue is handled in a more flexible manner that allows \textit{mixed initiative}, i.e.~a situation where the control over the dialogue moves back and forth between the agent and the user. Here, the agent maintains a so-called \textit{frame} containing several \textit{slots} that it attempts to fill by interacting with the user who may provide the information in any order (sometimes also filling more than one slot with a single sentence). Figure~\ref{fig:frame} shows (some of the) slots contained in a frame used in a flight-reservation system. 

\begin{figure}[t]
    \centering
        \begin{tabular}{l|l}
        \hline
         Slot    &  Question  \\
         \hline
         \texttt{DEPARTURE CITY} & \textit{Where are you traveling from?} \\ 
         \texttt{ARRIVAL CITY}   & \textit{Where are you going to?}  \\ 
         \texttt{DEPARTURE DATE} & \textit{What day would you like to travel?} \\ \texttt{DEPARTURE TIME} & \textit{What time would you like to leave?} \\
         \hline
        \end{tabular}
    \caption{An example of a frame for a flight-reservation system. The frame contains a set of slots, each with an associated question. Note that only some of the required slots are shown in this figure.}
    \label{fig:frame}
\end{figure}

In a single-domain frame-based system, the agent's first step is to determine the intent of the user. 
However, in the more common multi-domain systems, this step is preceded by a \textit{domain determination} step where the agent identifies the overall topic of discourse\cite{JurafskyMartin2009}. Intent detection can be implemented in the form of explicit, hand-written rules, or with the use of more sophisticated \textit{semantic grammars} that define a set of rules (involving phrases) that form a compact representation of many different variations of a sentence, allowing the CA to understand different user inputs with similar meaning. 

\begin{figure}[t]
    \centering
        \begin{tabular}{l}
        \hline
         Rule  \\
         \hline
         \texttt{QUERY}~$\rightarrow~\{$\textit{is there, do you have}$\}$ \\
\texttt{VEHICLE}~$\rightarrow~\{$\textit{a car (available), any car (available)}$\}$ \\
\texttt{PRICELIMIT}~$\rightarrow~\{$\textit{for less than, under, (priced) below}$\}$\\
\texttt{PRICE}~$\rightarrow~\{$\texttt{NUMBER}\textit{ (dollars)}$\}$ \\
\texttt{NUMBER}~$\rightarrow~\{$\textit{50, 100, 150, 200, 300, 500}$\}$ \\
\texttt{TIMERANGE}~$\rightarrow~\{$\textit{a day, per day, a week, per week}$\}$ \\
         \hline
        \end{tabular}
    \caption{An example of a simple semantic grammar for (some parts of) a car rental dialogue.}
    \label{fig:semanticgrammar}
\end{figure}

As a simple (and partial) example, consider a CA that handles car rentals. 
Here, a suitable semantic grammar may contain the rules shown in Fig.~\ref{fig:semanticgrammar}, so 
that the sentences \lq\lq\textit{Is there any car available for less than 100 dollars per day?}\rq\rq
and \lq\lq\textit{Do you have a car priced below 300 a week?}\rq\rq would both be identified as instances
of \texttt{QUERY - VEHICLE - PRICELIMIT - PRICE - TIMERANGE}. Semantic grammars can be built in
a hierarchical fashion (see, for example, how \texttt{NUMBER} is part of \texttt{PRICE} in the example just described), and can be tailored to represent much more complex features than in this simple example\cite{WardEtAl1992}.
The output of frame-based systems (i.e.~the NLG part in Figure~\ref{fig:pipelinemodel}) is 
typically defined using a set of predefined templates. Continuing with the car
rental example, such a template could take the form \lq\lq\textit{Yes, we have a} \texttt{\textlangle VEHICLETYPE\textrangle} \textit{
available at} \texttt{\textlangle NUMBER\textrangle} \textit{dollars per day}\rq\rq, where \texttt{\textlangle VEHICLETYPE\textrangle} and \texttt{\textlangle NUMBER\textrangle} would
then be replaced by the appropriate values for the situation at hand.

The frame-based approach works well in situations where the user is expected to provide
several pieces of information on a given topic. However, many dialogues involve
other aspects such as, for example, frequent switching back and forth between different domains
(contexts). 
Natural dialogue requires a more advanced and capable representation than that offered
by frame-based systems and typically involves identification of \textit{dialogue acts}
as well as \textit{dialogue state tracking}. Dialogue acts\cite{Bunt1994} provide a form
of categorization of an utterance, and represent an important first step in understanding
the utterance\cite{StolckeEtAl2000}. 
For example, an utterance of the form \lq\lq\textit{Show me all available
flights to London}\rq\rq is an instance of the \texttt{COMMAND} dialogue act, whereas the 
possible response \lq\lq\textit{OK, which day to you want to travel?}\rq\rq is an 
instance of \texttt{CLARIFY}, and so on. Dialogue act \textit{recognition} has been
approached in many different ways\cite{ChenEtAl2018}, reaching an accuracy of around 80-90\%,
depending on the data set used.
Dialogue state tracking, which is essentially the ability of the CA to keep track of context, 
has also been implemented in many ways, for example using a set of state variables
(as in MIT's JUPITER system\cite{ZueEtAl2000}) or the more general concept of 
\textit{information state} that underlies the TrindiKit and other dialogue
systems\cite{LarssonTraum2000}. Dialogue state tracking has also been the
subject of a series of competitions; see the work by Williams \textit{et al.}\cite{WilliamsEtAl2016} 
for a review.

For the remainder of this section, the systems under consideration reach a level of complexity such that
their interpretability is arguably diminished, yet they still
make of use high-level, interpretable primitives and are most
definitely far from being complete black boxes.
Thus, going beyond finite-state and frame-based systems, several methods 
have been developed that incorporate more
sophisticated models of the user's beliefs and goals.
These methods can generally be called \textit{model-based} methods\cite{HarmsEtAl2018} (even though
this broad category of methods also goes under other names, and involves somewhat confusing
overlaps between the different approaches).
Specifically, in \textit{plan-based} methods\cite{CohenPerrault1979, AllenEtAl2001},
such as Ravenclaw\cite{BohusRudnicky2009},
which are also known as belief-desire-intention (BDI) models, the CA explicitly
models the goals of the conversation and attempts to guide the user
towards those goals. 
\textit{Agent-based} methods\cite{Blaylock2005}
constitute another specific example. They involve aspects of both 
the plan-based and information state approaches, and
extend those notions to model dialogue as a form of cooperative 
problem-solving between agents (the user and the CA), where each agent
is capable of reasoning about its own beliefs and goals but also those
of the other agent. Thus, for example, this approach makes it possible for the CA not only to
respond to a question literally, but also to provide
additional information in anticipation of the user's goals\cite{McTear2002}.

Even though there is no fundamental obstacle to using a data-driven, machine learning
approach to systems of the kind described above, these methods are often referred
to as \textit{handcrafted} as this is the manner in which they have most often been
built. Moreover, these methods generally maintain a single, deterministic
description of the current state of the dialogue. There are also methods that
model dialogue in a probabilistic, statistical way and which can be seen as extensions
of the information-state approach mentioned above. The first steps towards such
systems model dialogue as a Markov decision process (MDP)\cite{LevinEtAl1998}, in
which there is a set of states ($S$), a set of actions ($A$), and
a matrix of transition probabilities ($P$) that models the probability
of moving from one given state to another when taking a specific action
$a \in A$. Any action $a$ resulting in state $s$ is associated with
an immediate reward $r(s,a)$. There is also a dialogue policy $\pi$,
specifying the action that the agent should take in any given situation.
The policy is optimized using reinforcement learning so as to maximize the
expected reward. 
Levin \textit{et al.}\cite{LevinEtAl2000} provide
an accessible introduction to this approach. The MDP approach, which assumes
that the state $s$ is observable, is unable to deal with the inevitable
uncertainty of dialogue, whereby the user's goal, the user's last
dialogue act, and the dialogue history, which together define
the dialogue state, cannot be known with absolute certainty\cite{YoungEtAl2007}.
Therefore, systems that take into account the inherent uncertainty 
in conversation (including ASR uncertainty) have been proposed in the
general framework of partially observable Markov decision processes (POMDPs).
Rather than representing a single state, a POMDP defines a distribution
over all possible dialogue states, referred to as the \textit{belief state}.
A difficult problem with MDPs and, to an even greater degree, POMDPs, is
the fact that their representation of dialogue states is very complex. 
This, in turn, results in huge state spaces and therefore computational 
intractability. 
Much work on POMDPs, therefore,
involves various ways of coping with such problems\cite{YoungEtAl2010}.
Young \textit{et al.} provide a review of POMDPs for dialogue\cite{YoungEtAl2013}.

\subsubsection{Black box dialogue systems}
\label{sect:blackbox}
Black box task-oriented agents are essentially dominated by the field of DNNs.
For these CAs, many of the aspects that were described before for DNN-based chatbots still apply.
For example, DNN-based task-oriented agents also represent words with embeddings, and rely on DNN architectures such as recurrent or convolutional ones\cite{WenEtAl2016}, as well as transformer-based ones\cite{BudzianowskiVuli2019,GouEtAl2020}.
There are other types of DNNs used for task-oriented systems, such as memory networks\cite{WestonEtAl2015,BordesEtAl2017} and graph convolutional networks\cite{KipfWelling2017,BanerjeeKhapra2019}. However, delving into their explanation is beyond the scope of this chapter.

Normally, the type of DNN used depends on what stage of the agent the DNN is responsible for. 
In fact, black box systems can often be framed in terms of the pipeline model (Figure~\ref{fig:pipelinemodel}), where NLU, NLG, dialogue policy, etc., are modeled somewhat separately. 
Even so, the DNNs responsible for realizing a black box task-oriented agent are connected to, and dependent of, one another, so as to allow the flow of information that makes end-to-end information processing possible.
An emerging trend in the field, as we mentioned before, is to make the agent less and less dependent on human design, and instead increase the responsibility of the DNNs to automatically learn how to solve the task from examples and interaction feedback.

Wen \textit{et al.}\cite{WenEtAl2016} proposed one of the first DNN-based task-oriented agents trained end-to-end, namely a CA for restaurant information retrieval.
To collect the specific data needed for training the system, the authors used a \emph{Wizard-of-Oz} approach\cite{Kelley1984}, scaled to a crowd-sourcing setting, to gather a large number of examples. 
In particular, one part of Amazon Mechanical Turk\cite{Crowston2012} workers were instructed to act as normal users asking information about restaurants and food. The other part, i.e., the \emph{wizards of Oz}, were asked to play the role of a perfect agent by replying to all inquiries using a table of information that was provided beforehand. 
Alternatively, since training from examples requires large corpora to be collected beforehand, reinforcement learning can be used\cite{LiuEtAl2017}. 

Some works attempt to improve the way these black box systems interface with the knowledge base.
Typically, the DNN that is responsible for the dialogue policy chooses the information retrieval query deemed to be most appropriate.
Dhingra \textit{et al.}\cite{DhingraEtAl2016} have shown, however, that one can include mechanisms to change the lookup operation as to use \emph{soft} queries that provide multiple degrees of truth, ultimately making it possible to obtain a richer signal that is useful to improve training.
Madotto \textit{et al.}\cite{MadottoEtAl2020} instead looked at improving the scalability of the systems to query large knowledge bases, by essentially having the DNNs assimilate information from the knowledge base within the network parameters at training time. This improves information retrieval speed because querying the external knowledge base normally takes longer than inferring the same information from within a DNN. However, this also means that some degree of re-training is needed when the information in the knowledge base is changed.

There are many more works on DNN-based task-oriented agents that deal with different aspects for improvement. 
For instance, a problem is that a DNN responsible for NLG can learn to rely on specific information retrieved from the knowledge base in order to generate meaningful responses, to such an extent that changes to the knowledge base can lead to unstable NLG. This problem is being
tackled in ongoing research aimed at decoupling NLG from the knowledge base\cite{RaghuEtAl2019,WuSocherEtAl2019}.
Another interesting aspect concerns incorporating training data from multiple domains, for situations where the data that are specific to the task at hand are scarce\cite{QinEtAl2020}.
Finally, we refer to the works by Bordes \textit{et al.}\cite{BordesEtAl2017}, Budzianowski \textit{et al.}\cite{BudzianowskiEtAl2018}, and  Rajendran \textit{et al.}\cite{RajendranEtAl2018} as examples where data sets useful for training and benchmarking this type of CA are presented.

\section{Evaluation of conversational agents}\label{sec:evaluation}
\begin{figure}
    \centering
    \includegraphics[width=\linewidth]{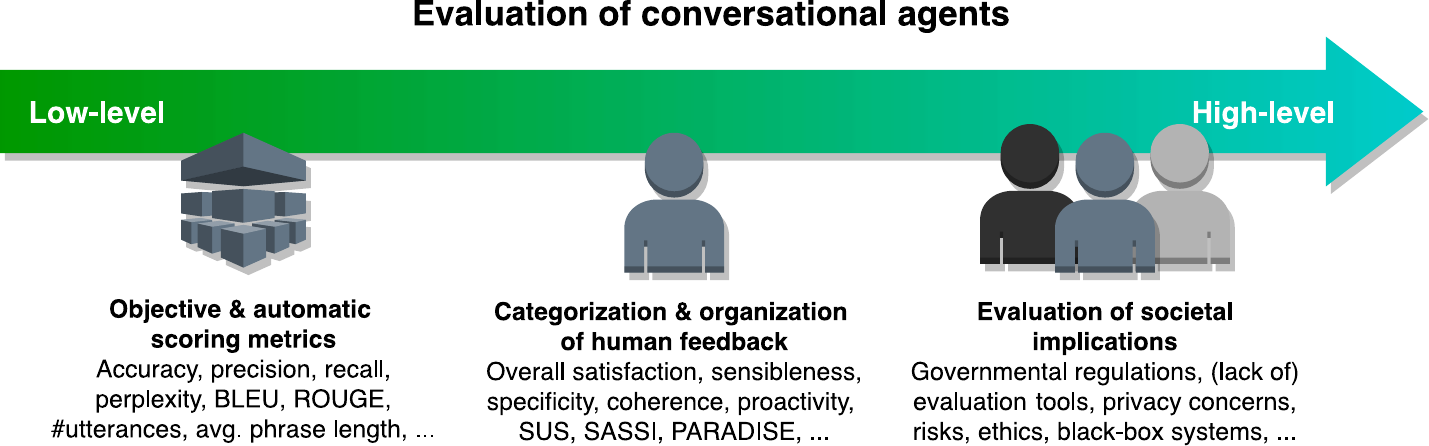}
    \caption{Schematic view of the different levels at which CAs can be evaluated, mirroring the organization of Section~\ref{sec:evaluation}.}
    \label{fig:evaluationCAs}
\end{figure}

Evaluation of CAs is very much an open topic of research. 
Current systems are still far from achieving seamless, human-like conversation capabilities\cite{Luger2016like,Levesque2017}. 
Understanding how best to measure such capabilities represents an important endeavor to advance the state-of-the-art. 
There are many different evaluation criteria, aimed at different levels of the human-agent interaction.
Firstly in this section, a brief introduction on low-level metrics used to evaluate general language processing is given.
Next, moving on to a higher-level of abstraction, evaluation metrics that address the quality of interaction with CAs, in a more general setting, and in more detail for the case of ECAs, are considered. Last but not least, evaluation systems that delve into broad implications and ethics are described.
Figure~\ref{fig:evaluationCAs} summarizes these different aspects of evaluation.

\subsection{Low-level language processing evaluation metrics}\label{sec:evaluationLowLevel}
On a low level, when developing the capabilities of a CA to process utterances, retrieve information, and generate language, traditional metrics for general pattern recognition like \emph{precision} and \emph{recall} represent useful metrics for evaluation. 
For example, for a CA employed in a psychiatric facility, one may wish the agent (or a component thereof) to 
assess whether
signs of depression transpire from the interaction with the patient: In such a case, maximizing recall is important.
In practice, traditional metrics like precision and recall are employed across several types of benchmark problems, e.g., to evaluate sentiment analysis\cite{SocherEtAl2013}, paraphrase identification\cite{DolanBrockett2005}, question answering\cite{RajpurkarEtAl2016}, reading comprehension\cite{ZhangEtAl2018}, but also gender bias assessment\cite{ZhaoEtAl2018}.
The \textit{general language understanding evaluation} (GLUE) benchmark\cite{WangEtAl2018}, and its successor SuperGLUE\cite{WangEtAl2019}, provide a carefully-picked collection of data sets and tools for evaluating natural language processing systems, and maintain a leaderboard of top-scoring systems.

Alongside traditional scoring metrics, the field of natural language processing has produced additional metrics that are useful to evaluate CAs in that they focus on words and $n$-grams (sequences of words), and are thus more directly related to language.
A relevant example of this is the \textit{bilingual evaluation understudy} (BLEU) scoring metric. 
Albeit originally developed to evaluate the quality of machine translation\cite{PapineniEtAl2002}, BLEU has since been adopted in many other tasks, several of which are of relevance for CAs, e.g., summarization and question answering. 
For the sake of brevity, we will only describe how BLEU works at a high level. 
In short, BLEU can be seen as related to precision and is computed by considering how many $n$-grams in a candidate sentence (e.g., produced by a CA) match $n$-grams in reference sentences (e.g., high-quality sentences produced by humans). The $n$-grams considered are normally of length $1$ to $4$. Matches for longer $n$-grams are weighted exponentially more than matches for smaller $n$-grams.
In addition, BLEU includes a penalty term to penalize candidate sentences that are shorter than reference sentences.
We refer the reader interested in the details of BLEU to the seminal work by Panineni \textit{et al.}\cite{PapineniEtAl2002}.
A metric similar to BLEU, but with the difference that it focuses on recall rather than precision, is \textit{recall-oriented understudy for gisting evaluation} (ROUGE), which is reviewed in the seminal works by Lin~\textit{et al.}\cite{Lin2004,LinEtAl2004}.
    
\subsection{Evaluating the quality of the interaction}\label{sec:evaluationHuman}
The evaluation of the quality of interaction requires human judgment. 
In this context, researchers attempt to find the aspects that define a good interaction with an agent, in order to obtain a better understanding of how CAs can be built successfully. 
We divide this section into one general part on conversational capabilities, and one part that is focused on embodiment.

\subsubsection{General CAs}
The work by Adiwardana \textit{et al.} of Google Brain proposes to categorize human evaluation into \emph{sensibleness} and \emph{specificity}\cite{AdiwardanaEtAl2020}. 
Evaluating sensibleness means to assess whether the agent is able to give responses that make sense in context. 
Evaluating specificity, instead, involves determining whether the agent is effectively capable of answering with specific information, as opposed to providing vague and empty responses, a strategy often taken to avoid making mistakes. 
Interestingly, the authors found that the average of sensibleness and specificity reported by users correlates well with \emph{perplexity}, a low-level metric for language models that represents the degree of uncertainty in token prediction.

The Alexa Prize competition represents an important stage for the evaluation of CAs\cite{Amazon2020}. 
To rank the contestants, Venkatesh \textit{et al.}~defined several aspects, some of which require human assessment, and some that are automatic\cite{VenkateshEtAl2018}.
For example, one aspect is \emph{conversational user experience}, which is collected as the overall rating (in the range $[1,5]$) from users that were given the task of interacting with the CAs during the competition. Another aspect, \emph{coherence}, was set to be the number of sensible responses over the total number of responses, with humans annotating sensible answers. 
Other aspects were evaluated with simple metrics, e.g., \emph{engagement} as the number of utterances, and \emph{topical breadth} and \emph{topical depth} as the number of topics touched in the conversation (identified with a vocabulary of keywords) and the number of consecutive utterances on a certain topic, respectively. Venkatesh \textit{et al.} also attempted to find a correlation between human judgment and automatic metrics, but only small correlations were found.

The recent survey by Chaves and Gerrosa\cite{ChavesGerosa2020} reports on further research endeavors
to elucidate the aspects that account for the perception of successful and engaging interactions with CAs.
These aspects include, for example, \emph{conscientiousness}: how aware of and attentive to the context the CA appears to be; \emph{communicability}: how transparent the CA's interaction capabilities are, e.g., for the user to know how to best query the agent\cite{HillEtAl2015}; \emph{damage control}: how capable the CA is to recover from failure, handle unknown concepts; etc\cite{LiEtAl2018}.
Some apparent qualities can actually be perceived to be negative, when excessive.
CAs that take too much initiative constitute a clear example, as they can be perceived as intrusive or frustrating, or even give the feeling of attempting to exercise control over the user\cite{Duijvelshoff2017,TallynEtAl2018}.

Aspects of the perception of the interaction are collected using standardized questionnaires (involving, for example, the Likert scale\cite{Likert1932}), which can be relatively general and can be applied beyond CAs, like the \textit{system usability scale} (SUS)\cite{Brooke1996}, or specific to the human-CA interaction, including particular focuses such as evaluation of speech recognition, as done by the \textit{subjective assessment of speech system interfaces} (SASSI)\cite{HoneGraham2000}.
A notable organizational framework of user feedback that applies well to task-oriented CAs is the \textit{paradigm for dialogue system evaluation} (PARADISE)\cite{WalkerEtAl1997}. 
PARADISE links the overall human-provided rating about the quality of interaction to the agent's likelihood of success in completing a task and the cost to carry out the interaction. 
Task success is based upon pre-deciding what the most important building blocks of the interaction are, in the form of attribute-value pairs (e.g., an attribute could be \emph{departing station}, and respective values could be \emph{Amsterdam}, \emph{Berlin}, \emph{Paris}, \dots). 
With this organization of the information, confusion matrices can be created, useful for determining whether the CA returns the correct values for interactions regarding certain attributes. 
The cost of the interactions, i.e., how much effort the user needs to invest to reach the goal, consists of counting how many utterances are required in total, and how many are needed to correct a misunderstanding. 
Finally, a scoring metric is obtained by a linear regression of the perceived satisfaction as a function of task success statistics and interaction costs.

Recently, in an effort to provide consistent evaluations and to automate human involvement, Sedoc \textit{et al.}~proposed ChatEval\cite{SedocEtAl2019}, a web application aimed at standardizing comparisons across CAs. 
To use ChatEval, one uploads (the description and) the responses of a CA to a provided set of prompts. 
As soon as they are uploaded, the responses are evaluated automatically with low-level metrics such as average response length, BLEU score, and cosine similarity over sentence embeddings between the agent's
response and the ground-truth response\cite{LiuEtAl2016}. 
Optionally, by paying a fee one can make ChatEval automatically start a human-annotation task on Amazon Mechanical Turk, where users evaluate the responses of the agent in a standardized way.

\subsubsection{Evaluation of embodiment}
When embodiment comes into play, alongside aspects that are general to the interaction with CAs such as conversational capabilities in both chat and task-oriented settings, or handling of intentional malevolent probing, ECAs can be evaluated in terms of \emph{visual look} and \emph{appearance personalization options}\cite{Kuligowska2015}.
Another aspect of evaluation regards whether the (E)CA exhibits (interesting) \emph{personality traits}.
ECAs are arguably advantaged in this context compared to non-embodied CAs, because they can enrich their communication with body language and facial expressions.

Since ECAs can leverage nonverbal interaction, they can engage the user more deeply, at an emotional level. 
In this setting, emotional ECAs can be evaluated by comparison with non-emotional ones\cite{MeiraCanuto2015}, in terms of aspects such as \emph{likeability}, \emph{warmth}, \emph{enjoyment}, \emph{trust}, \emph{naturalness of emotional reactions}, \emph{believability}, and so on.

Evaluation of embodiment can also be carried out in terms of opposition with the absence of embodiment, or in terms of physical embodiment vs.~virtual embodiment, the latter type of comparison being more popular.
In one such study\cite{LeeEtAl2006}, it was found that people evaluate (at different levels) the interaction with a physical ECA (specifically, Sony's Aibo\cite{Aibo2020}) and a prototype robot by Samsung, called April), more positively than the interaction with a virtual version of the same ECA. 
A similar study has been carried out using a physical and virtual ECA to interact during chess games\cite{PereiraEtAl2008}.
At the same time, other studies have instead recorded that physical ECAs are no better than virtual ECAs. 
For example, it was found that whether a task-oriented ECA is physical or virtual results in no statistical differences in two very different tasks: persuading the user to consume healthier food, and helping to solve the Towers of Hanoi puzzle\cite{HoffmannKramer2013}.
Last but not least, as mentioned before in Section~\ref{sec:input-output modalities}, an \emph{excess} of embodiment (particularly so when embodiment attempts to be too human-like), can actually worsen the quality of the perceived interaction. 
Aside from the uncanny valley hypothesis\cite{MoriEtAl2012,CiechanowskiEtAl2019}, another valid reason for this is that more sophisticated embodiments can lead to bigger expectations on conversational capabilities that leave the user more disappointed when those expectations are not met\cite{MimounEtAl2012}.

Much uncertainty about the benefits of using physical or virtual ECAs is still present, since 
both the ECAs involved and the methods used for their evaluations vary wildly in the literature,
from adopting questionnaires on different aspects of the interaction to measuring how long it takes to complete a puzzle.
For further reading on the evaluation of ECAs, we refer to Granstr\"om and House\cite{GranstromHouse2007}, and Weiss \textit{et al.}\cite{WeissEtAl2015}.

\subsection{Evaluating societal implications}\label{sec:evaluationSocietalImplications}
With CAs becoming ever more advanced and widespread, it is becoming increasingly clear that their societal impact must be evaluated carefully before they are deployed.
In fact, failures of CAs from a societal and ethical standpoint have already being recorded. 
Microsoft's chatbot \emph{Tay} for Twitter is an infamous example\cite{NeffNagy2016}:~The agent's capability to learn and mimic user behavior was exploited by Twitter trolls, who taught the agent to produce offensive tweets, forcing a shutdown of the agent.
Moreover, it is not hard to imagine that a lack of regulations and evaluation criteria for the safe deployment of CAs in sensitive societal sectors can lead to a number of unwanted consequences.
While some general regulations might exist\cite{GoodmanFlaxman2017}, for many societal sectors the evaluation tools required to ensure safe and ethical use of CAs are still largely missing.

Healthcare is one among several sectors where the adoption of CAs could bring many benefits, e.g., by helping patients to access care during a pandemic\cite{BhartiEtAl2020} or support pediatric cancer patients who must remain in isolation during part of their treatment\cite{LigthartEtAl2019}. 
However, many risks exist as well\cite{Luxton2020}.
For example, safeguarding patient privacy is crucial: It must be ensured that a CA does not leak sensitive patient information. 
Ethical and safety implications regarding the circumstances and conditions under which a CA
is allowed to operate in a clinic must also be carefully assessed. 
Examples of questions in this domain are: 
\lq\lq\emph{In what context is it ethical to have patients interacting with a CA rather than a human?}\rq\rq;
\lq\lq\emph{Is informing about health complications a context where an agent should be allowed to operate?}\rq\rq;
\lq\lq\emph{Is it safe to use CAs for patients who suffer from psychiatric disorders?}\rq\rq;
\lq\lq\emph{Would it be acceptable to introduce disparities in care whereby wealthier patients can immediately access human doctors, while the others must converse with agents first?}\rq\rq.
Evaluation tools to answer these questions have yet to be developed.

Last but not least, it is essential to note that the use of black box CAs comes with additional risks,
especially in cases where such CAs are involved in high-stakes decisions, where
one can instead argue in favor of more transparent and accountable systems\cite{Rudin2019}.
The field of \textit{interpretable AI} involves methods for developing such inherently transparent systems. The closely related, and perhaps more general, notion of \textit{explainable AI} (XAI)\cite{AdadiBerrada2018,GuidottiEtAl2018} considers (post-hoc) approaches for attempting to explain the decisions taken by black box systems\cite{Rudin2019}
as well as methods for improving their level of explainability\cite{AngelovSoares2020}.

With transparent systems originating from interpretable AI, a sufficient understanding of
the system's functionality to guarantee safe use can generally be achieved. For black box
CAs, however, one should be careful when applying methods for explaining their decision-making
to the specific case of evaluating their \textit{safety}. 
This is so, since explanations for black box systems can, in general, only be specific (also referred to 
as \emph{local}\cite{AdadiBerrada2018}) to the particular event under study.  
In other words, given two events that are deemed to be similar and for which it is expected that the CA should react similarly, this might not happen\cite{Rudin2019}. 
Furthermore, the possibility to understand the workings of agent components derived using machine learning 
is key to spotting potential biases hidden in the (massive amounts of) data\cite{AdadiBerrada2018,GuidottiEtAl2018} used when training such components.
With black box systems, it is simply harder to pinpoint potential damaging biases before the system has been released and has already caused damage\cite{Rudin2019,Small2019}.

\section{Notable conversational agents}
\label{sect:history}
To some degree, this section can be considered as a historical review of CAs, but it should be observed that the set of CAs that have gained widespread attention is very much dominated by chatbots rather than task-oriented agents: even though some task-oriented systems are mentioned towards the end of the section, for a complete historical review the reader is also advised to consider the various systems for task-oriented dialogue presented in Subsection~\ref{subsect:taskoriented} above. Moreover, this section is focused on dialogue capabilities rather than than the considerable and concurrent advances in, say, speech recognition and the visual representation of CAs (embodiment).

The origin of CAs can be traced back to the dawn of the computer age when, in 1950, Turing reflected on the question of whether a machine can think\cite{Turing1950}. Noting that this question is very difficult to answer, Turing introduced \textit{the imitation game}, which today goes under the name \textit{the Turing test}.
In the imitation game, a human interrogator ($C$) interacts with two other participants, a human ($A$) and a machine ($B$). The participants are not visible to each other, so that the only way that $C$ can interact with $A$ and $B$ is via textual conversation. $B$ is said to have passed the test if $C$ is unable to determine that it is indeed the machine.

Passing the Turing test has been an important goal in CA research (see also below) even though, as a measure of machine intelligence, the test itself is also controversial; objections include, for example, suggestions that the test does not cover all aspects of intelligence, or that successful imitation does not (necessarily) imply actual, conscious thinking\footnote{It should be noted, in fairness, that Turing did address many of those objections in his paper, and that he did not intend his test to be a measure of machine consciousness.}.

As computer technology became more advanced, the following decade saw the introduction of the first CA, namely the chatbot ELIZA\cite{Weizenbaum1966} in 1966; see also Subsection~\ref{subsubsect:patternbased} above. The most famous incarnation of ELIZA, called DOCTOR, was implemented with the intention of imitating a Rogerian psychoanalyst. Despite its relative simplicity compared to modern CAs, ELIZA was very successful; some users reportedly engaged in deep conversation with ELIZA, believing that it was an actual human, or at least acting as though they held such a belief.

In 1972, ELIZA was followed by the PARRY chatbot that, to a great degree,
represents its opposite: Instead of representing a medical professional, PARRY was
written to imitate a paranoid schizophrenic, and was used as a training
tool for psychiatrists to learn how to communicate with patients suffering from this affliction. 
The semblance of intelligence exhibited by both ELIZA and PARRY can be derived, to a great degree, from their ability to deflect the
user's input, without actually giving a concrete, definitive reply to
the user. Predictably, ELIZA has actually met PARRY in several conversations\cite{ElizaParry1972}, which took place in 1972 over the ARPANet, the predecessor to the internet.

At this point, it is relevant to mention the Loebner prize, which was instituted in the early 1990s. In the associated competition, which is an annual event since 1991, the entrants participate in a Turing test. A 100,000 USD one-time prize (yet to be awarded) is offered for a CA that can interact with (human) evaluators in such a way that it is deemed indistinguishable from a real human in terms of its conversational abilities. There are also several smaller prizes. The Loebner competition is controversial and has been criticized in several different ways, for example on the grounds that it tends to favor deception rather than true intelligence. Nevertheless, the list of successful entrants does offer some insight into the progress of CA development.

A.L.I.C.E\cite{Wallace2009}, a somewhat contrived abbreviation of Artificial Linguistic Internet Computer Entity, was a pattern-based chatbot built using AIML (see Section~\ref{subsubsect:patternbased}). Its first version was released in 1995, followed by additional versions a few years later. Versions of A.L.I.C.E. won the Loebner competition several times in the early 2000s. 

Two other notable chatbots are Jabberwacky and its successor Cleverbot.
Jabberwacky was implemented in the 1980s, and was released on the internet in 1997,
whereas Cleverbot appeared in 2006. These chatbots are based on information retrieval\cite{Carpenter2020} and improve their capabilities (over time) automatically
from conversation logs: every interaction between Cleverbot and a human is stored, and
can then be accessed by the chatbot in future conversations. Versions of Jabberwacky won the Loebner competition in 2005 and 2006.
Cleverbot rose to fame when, in 2011, it participated in a Turing test (different from the Loebner competition) in India, and was deemed 59.3\% human, which can be compared with the average score of 63.3\% for \textit{human} participants. Another chatbot for which similar performance has been reported is Eugene Goostman. 
In two Turing tests\cite{ShahEtAl2016}, one in 2012 marking the centenary of the birth of Alan Turing, and one in 2014, organized in Cambridge on the occasion of the 60$^\textrm{th}$ anniversary of his death, this chatbot managed to convince 29\% (2012) and 33\% (2014) of the evaluators that it was human. Claims regarding the intellectual capabilities of these chatbots have also been widely criticized.

Another recent AIML-based chatbot is Mitsuku\cite{Mitsuku2020}, which features interactive learning and can perhaps be considered as a representative of the current state-of-the-art in pattern-based chatbot technology. Mitsuku has won the Loebner competition multiple times (more than any other participant) in the 2010s.
Implemented in a scripting language called ChatScript\cite{Wilcox2020}, the chatbot Rose and its two predecessors Suzette and Rosette have also won the Loebner competition in the 2010s.

The 2010s also saw the introduction of CAs used as personal assistants on mobile devices, starting with the launch of SIRI (for Apple's iPhone) in 2011, and then rapidly followed by Google Now in 2012, as well as Amazon's Alexa and Microsoft's Cortana a few years later\footnote{SmarterChild, a CA released in the early 2000s on instant messenger networks, can perhaps be seen as a precursor to SIRI.}. These CAs, which also generally feature advanced speech recognition, combine chatbot functionality (for everyday conversation) with task-oriented capabilities (for answering specific queries). In this context, it is relevant to mention again the Alexa prize\cite{Amazon2020}; introduced in 2016, this contest does not focus on the Turing test. Instead, the aim is to generate a CA that is able to converse \lq\lq coherently and engagingly\rq\rq on a wide range of current topics, with a grand challenge involving a 20-minute conversation of that kind\cite{RamEtAl2018}.

In 2020, Google Brain presented Meena\cite{AdiwardanaEtAl2020}, a new DNN-based generative chatbot the architecture of which was optimized with evolutionary algorithms. With human evaluation recorded in percentage and based on the average of sensibleness and specificity (see Section~\ref{sec:evaluationHuman}), Meena was found to outperform chatbots such as XiaoIce, DialoGPT, Mitsuku, and Cleverbot, with a gain of 50\% compared to XiaoIce and 25\% compared to Cleverbot (the best after Meena).

\section{Applications}
\label{sect:applications}
This section provides a review of recent applications of CAs, focusing mainly on 
task-oriented agents. We have attempted to organize applications into
a few main subtopics, followed by a final subsection on other applications.
However, we make no claims regarding the completeness of the survey below; in
this rapidly evolving field new interesting applications appear continuously.
The description below is centered on results from scientific research. There are
also several commercial products that are relevant, for example general-purpose
CAs such as Apple's Siri, Amazon's Alexa, and Microsoft's Cortana, and
CA development tools such as Google's DialogFlow, Microsoft's LUIS, 
IBM's Watson, and Amazon's Lex. Those products will not be considered
further here, however.

\subsection{Health and well-being}
Arguably one of the most promising application areas, health and
well-being is a central topic of much CA research\cite{LaranjoEtAl2018,MontenegroEtAl2019}.
This application area is not without controversy, however, due to ethical
considerations such as safety and respect for privacy\cite{Luxton2020}.

With the continuing rise in human life expectancy, especially in developed countries, 
the fraction of elderly people in the population is expected to rise dramatically
over the coming decades, an undoubtedly positive development but also one that will 
exacerbate an already strained situation for the healthcare systems in many
countries. Thus, a very active area of research is the study of 
CAs (and, especially, ECAs) for \textit{some} (non-critical) tasks in elderly care.
Examples include CAs that monitor medicine intake\cite{Fadhil2018},
interact with patients regarding their state of health (e.g.,~during cancer treatment)\cite{PiauEtAl2019}, 
or provide assistance and companionship\cite{SakaiEtAl2012,YaghoubzadehEtAl2013,NikitinaEtAl2018},
a case where also social robots, such as Paro by Shibata~\textit{et al.}\cite{ShibataEtAl2001} (see also Hung et al.\cite{HungEtAl2019}), play a role.

Mental health problems affect a large portion of the worldwide population.
Combined with a shortage of psychiatrists, especially in low-income
regions, these diseases present a major challenge\cite{GaffneyEtAl2019}.
CAs are increasingly being applied in the treatment of mental health
problems\cite{VaidyamEtAl2019,GaffneyEtAl2019,BendigEtAl2019}, generally with
positive results, even though most studies
presented so far have been rather preliminary and exploratory in their nature,
only rarely involving full, randomized controlled trials.
Moreover, issues beyond the CA functionality and performance, such as legal and 
ethical considerations, must also be addressed\cite{Luxton2020,VaidyamEtAl2019}, 
carefully and thoroughly, before CAs can be applied fully in
the treatment of mental health problems (or, indeed, in any form of healthcare).
Several tasks can be envisioned for CAs applied to mental health problems.
A prominent example is intervention in cases of anxiety and depression, using
CAs such as Woebot\cite{Woebot2020} that applies cognitive behavioral therapy (CBT). Another important application is the treatment of post-traumatic
stress disorder (PTSD), where the anonymity offered by a CA interaction
(as opposed to an interaction with a human interviewer) may offer benefits\cite{LucasEtAl2017}.

CAs can also be used in lifestyle coaching, discouraging
harmful practices such as smoking\cite{PerskiEtAl2019} or drug abuse\cite{BhaktaEtAl2014},
and promoting healthy habits and lifestyle choices involving, for example, diet or
physical exercise\cite{FadhilGabrielli2017}. An example of such a CA is Ally, presented by Kramer~\textit{et al.}\cite{KramerEtAl2020}, which was shown to increase physical activity 
measured using step counts. A related application area is self-management of chronic diseases, 
such as asthma and diabetes. In the case of asthma, the kBot CA was used for helping
young asthma patients manage their condition\cite{KadariyaEtAl2019}. 
Similarly, Gong~\textit{et al.}\cite{GongEtAl2020} proposed Laura, which was
tested in a randomized controlled trial and was shown to provide significant benefits.

Another important application, where spoken dialogue is crucial, is to provide assistance
to people who are visually impaired with the aim of improving social
inclusion as well as accessibility to various services\cite{FelixEtAl2018,BighamEtAl2008}.

CAs may also come to play an important role in assisting medical professionals, for
example in scheduling appointments, providing information to patients, as well as
in dictation. A survey of 100 physicians showed that a large majority believe that
CAs could be useful in administrative tasks, but less so in clinical tasks\cite{PalanicaEtAl2019}.
Still, some CAs, such as Mandy by Ni~\textit{et al.}\cite{NiEtAl2017}, aim to provide assistance to
physicians in primary care, for example in basic interaction with patients regarding their condition.

\subsection{Education} 
Education is another field where CAs have been applied widely.
Most of the educational applications reported below rely on interpretable CAs (e.g., based on AIML\cite{Roos2018}), however black box CAs have also been investigated (e.g., using seq2seq\cite{PalasundramEtAl2019}).
A natural application of CAs in the education domain is to provide a scalable approach to
knowledge dissemination.
Massive open online courses are a clear example, since a one-to-one interaction between teacher and student is not feasible in those cases. In such settings, the course quality can be improved by setting up CAs that handle requests from hundreds or thousands of students at the same time\cite{WinklerSoellner2018}.
For instance, Bayne \textit{et al.} proposed Teacherbot\cite{Bayne2015}. This pattern-based CA was created to provide teaching support over Twitter for a massive open online course that enrolled tens of thousands of students.
Similarly, a CA in the form of a messaging app was created to provide teaching assistance for a large course at the Hong Kong University of Science and Technology\cite{GondaChu2019}.

Even though CAs can improve the dissemination of education, their effectiveness in terms of learning outcomes and engagement capability strongly depend on the specific way in which they are implemented, and the context in which they are proposed. 
For example, AutoTutor was found to compare favorably when contrasted with classic textbook reading in the study of physics\cite{GraesserEtAl2004}.
On the other hand, the AIML-based CA Piagetbot (a successor of earlier work, Freudbot\cite{HellerEtAl2005}) was not more successful than textbook reading when administered to psychology students\cite{HellerProctor2007}.
More recently, Winkler \textit{et al.} proposed Sara\cite{WinklerEtAl2020}, a CA that interacts with students during video lectures on programming. Interestingly, equipping Sara with voice recognition and structuring the dialogue into sub-dialogues to induce scaffolding, were found to be key mechanisms to improve knowledge retention and the ability to transfer concepts.
Beyond providing an active teaching support, CAs have also been used to improve the way education is conducted, e.g., as an interface for course evaluation\cite{WambsganssEtAl2020}, and related logistics aspects, e.g., by providing information about the campus\cite{GhoseBarua2013,RanoliyaEtAl2017,DibitontoEtAl2018}.

Linking back to and connecting with the previous section on health and well-being, educational CAs have also been applied to provide learning support for visually-impaired adults\cite{KumarEtAl2016} and deaf children\cite{ColeEtAl1999}.
Moreover, a particularly sensitive context of care where CAs are being investigated is the education of children with autism. 
For example, an AIML-based virtual ECA called Thinking Head was adopted to teach children with autism (6 to 15 years old) about conversation skills and dealing with bullying\cite{MilneEtAl2010}.
Similarly, the ECA Rachel\cite{MowerEtAl2011} was used to teach emotional behavior.
More recently, ECAs have also been used to help teenagers with autism learn about nonverbal social interaction\cite{TanakaEtAl2017,AliEtAl2020}. Alongside speech recognition, these ECAs include facial features tracking, in order to provide cues on smiling, maintaining eye contact, and other nonverbal aspects of social interaction.

For more reading on this topic, we refer to the works by Johnson \textit{et al.}\cite{JohnsonEtAl2000}, Kerry \textit{et al.}\cite{KerryEtAl2008}, Roos\cite{Roos2018}, Veletsianos and Russell\cite{VeletsianosRussell2014}, and Winkler and S\"ollner\cite{WinklerSoellner2018}.

\subsection{Business applications} 
The rise of e-commerce has already transformed the business landscape and it is process that, in all likelihood, will continue
for many years to come. Currently, e-commerce is growing at a rate of 15-20\% per year worldwide, and the total value
of e-commerce transactions is more than 4 trillion US dollars. 
Alongside this trend, many companies are also deploying CAs in their sales and customer service. It has
been estimated that CA technology will be able to answer up to 80\% of users' questions, and provide savings
worth billions of dollars\cite{AdamEtAl2020}. 

Research in this field is often centered on customer satisfaction. For example Chung \textit{et al.}\cite{ChungEtAl2018} considered 
this issue in relation to customer service CAs for luxury brands, whereas Feine \textit{et al.}\cite{FeineEtAl2019} studied how
the customer experience can be assessed using sentiment analysis, concluding that
automated sentiment analysis can act as a proxy for direct customer feedback.
An important finding that must be taken into account when developing customer service CAs is that customers generally prefer systems that provide a quick and efficient solution to their problem\cite{DixonEtAl2010},
and that embodiment does not always improve users' perception of the interaction\cite{MimounEtAl2012}.
Another useful design principle is to use a tiered approach, where a CA can refer to a human
customer service agent in cases where it is unable to fulfill the customer's request\cite{CranshawEtAl2017}. As in the case of other applications, the development of CAs for business applications is facilitated by the advent of relevant data sets, in this case involving customer service conversations\cite{HardalovEtAl2018}.

\subsection{Tourism and culture} 
CAs are becoming increasingly popular means to improve and promote tourism.
In this context, a verbal interaction with a CA can be useful to provide guidance at the airport\cite{Kasinathan2020}, assist in booking recommendations\cite{NicaEtAl2018,SanoEtAl2018}, and to adapt tour routes\cite{CasilloEtAl2020}.
To improve portability, oftentimes these CAs are implemented as text-based mobile apps\cite{NiculescuEtAl2014,CasilloEtAl2020,AlotaibiEtAl2020}, sometimes making use of social media platforms\cite{Kyaw2018}.

Several CAs have been designed specifically to provide guidance and entertainment in museums and exhibits.
Gustafson \textit{et al.}~built August, an ECA consisting of a floating virtual head, and showcased it at the Stockholm Cultural Centre\cite{GustafsonEtAl1999}; Cassell \textit{et al.}~set up MACK, a robot-like ECA, at the MIT Media Lab\cite{CassellEtAl2002}; Traum \textit{et al.}~proposed Ada and Grace, two more ECAs, employed at the Museum of Science of Boston\cite{TraumEtAl2012}; and Kopp \textit{et al.}~studied how the visitors of the Heinz Nixdorf Museums Forum interacted with Max, a full-body human-like virtual ECA\cite{KoppEtAl2005}.
Seven years after the introduction of Max, Pfeiffer \textit{et al.}~compiled a paper listing the lessons learned from using the ECA\cite{PfeifferEtAl2011}. 
Beyond queries about the venue, tourists were found to ask about Max's personal background (e.g., \lq\lq\textit{Where are you from?}\rq\rq) and physical attributes (e.g., \lq\lq\emph{How tall are you?}\rq\rq), but also often insult and provoke the agent, overall testing the human-likeness of the CA's reactions.
For the interested reader, the short paper by Schaffer \textit{et al.}\cite{SchafferEtAl2018} describes key steps towards the development of a technical solution to deploy CAs in different museums.

While the previous works were targeted on a specific museum or venue, CulturERICA can converse about cross-museum, European cultural heritage\cite{MachidonEtAl2020}.
Like in the case of tourism, many museum-guide CAs are implemented as mobile apps, or as social media messaging platform accounts, for mobility\cite{VassosEtAl2016,GaiaEtAl2019}. 

For the reader interested in a similar application, namely the use of CAs in libraries, we refer to the works by Rubin \textit{et al.}\cite{RubinEtAl2010} and Vincze\cite{Vincze2017}.

\subsection{Other applications}

There are many more and wildly different applications where CAs can be beneficial.
For example, CAs are being investigated to aid legal information access. Applications of this kind include legal guidance for couple separation and child custody\cite{AgredaEtAl2013}, as well as immigration and banking\cite{QueudotEtAl2020}.
DNN-based CAs are also being investigated to summarize salient information from legal documents in order to speed up the legal process\cite{ShubhashriEtAl2018}.

Research is also being conducted regarding the role of CAs in vehicles, and especially so in the realm of self-driving cars\cite{Lugano2017}.
Since vehicles are becoming increasingly autonomous, CAs can act as driver assistants in several ways, from improving passenger comfort by chatting about a wide range of activities, possibly unrelated to the ride (e.g., meetings, events), to assessing whether the driver is fit to drive, and even commanding the driver to take back vehicle control in case of danger\cite{OkurEtAl2020,deSalisEtAl2020}.
   
Games are another application of relevance for CAs.  
Beyond entertainment, games are often built to teach and train (so-called \emph{serious games}): In this setting, CAs are used to improve engagement in disparate applications, e.g., guiding children in games about healthy dietary habits\cite{FadhilVillafiorita2017}, teaching teenagers about privacy\cite{BergerEtAl2019}, and training business people as well as police personnel to react appropriately to conflicts and emergencies\cite{OthlinghausEtAl2020}.
Researchers have also investigated the effect of using CAs to help gaming communities grow and bond, by having the community members conversing with a CA in a video game streaming platform\cite{SeeringEtAl2020}.
Furthermore, CAs themselves represent a gamification element that can be of interest for the media: The BBC is exploring the use of CAs as a less formal and more fun way of reaching a younger audience\cite{JonesJones2019}.

Finally, CAs are often proposed across different domains to act as recommender systems\cite{JannachEtAl2020}.
Other than in business and customer service applications, recommender system-like CAs have been developed to recommend music tracks\cite{JinEtAl2019}, solutions for sustainable energy management\cite{GnewuchEtAl2018}, and also food recipes: Foodie Fooderson, for example, converses with the user to learn his or her preferences and then uses this information to recommend recipes that are healthy and tasty at the same time\cite{AngaraEtAl2017}.

\section{Future directions}
\label{sect:future}
Over the next few years, many new CAs will be deployed,
as much by necessity as by choice. 
For example, the trend towards online retailing is an important factor driving the development of CAs for customer 
service; similarly, the ongoing demographic shift in which an increasing fraction of
the population becomes elderly implies a need for technological 
approaches in healthcare, including, but not limited to, CAs;
furthermore, the advent of self-driving vehicles will also favor
the development of new types of CAs, for example those that operate
essentially as butlers during the trip, providing information and entertainment.
The further development of CAs may also lead to a strong shift
in the manner in which we interact with AI systems, such that
current technology (e.g.,~smartphones) may be replaced by
immersive technologies based on augmented or virtual reality.
Just like smartphones made interaction with screens more natural (touch as opposed to keyboards and smart pencils), we can expect that advanced, human-like CAs will lead to the development 
of new devices, where looking at a screen becomes secondary, and
natural conversation is the primary mode of interaction.

While the current strong focus on black box systems is likely to
persist for some time, we also predict that the widespread
use of CA technology, especially for task-oriented agents operating 
in areas that involve high-stakes decision-making as well as 
issues related to privacy, where accountability and interpretability are crucial,
will eventually force a shift towards more transparent
CA technologies, in which the decision-making can be
followed, explained and, when needed, corrected. 
This shift will be driven not only by ethical considerations
but also by legal ones\cite{JobinEtAl2019}. 

In the case of chatbots (e.g.,~Tay; see Subsection~\ref{sec:evaluationSocietalImplications}), 
we have already witnessed that 
irresponsible use of black box systems comes with risks, as
these systems are trained based on large corpora of (dialogue) data
that may be ridden with inherent biases towards the current majority view,
something that could put minority groups at a further disadvantage\cite{BenderEtAl2021}.
This raises an interesting point since one of the supposed advantages
with black box systems is that they reduce the need for handcrafting,
yet they may instead require more effort being spent in manually 
curating the data sets on which they rely.

As always, technology itself is neither good nor evil, but it can 
be used in either way; there is a growing fear that CAs might be used unethically, for example in gathering private conversational data for 
use in, say, mass surveillance and control. 
Moreover, as
technology already exists for generating so-called deep fakes 
(such as fake text, 
speech, or videos, 
wrongfully attributed to a specific person or group~\cite{AdelaniEtAl2020,BartoliEtAl2016,BartoliMedvet2020,NguyenEtAl2020,Westerlund2019}), 
CA technology could exacerbate the problem by allowing the development and deployment of
legions of fake personas that are indistinguishable from real persons 
and that may swarm social networks with malicious intent.
Even in cases where intentions are good, the use of CAs may
be controversial. For example, affective agents
may promote unhealthy or unethical bonding between humans
and machines. The research community and policymakers have a strong collective 
responsibility to prevent unethical uses of CA technology.
Tools must be developed for evaluating not only the functionality 
but also the \textit{safety} and \textit{societal impact} of CAs. 
As mentioned in Subsection~\ref{sec:evaluationSocietalImplications},
this is a nascent field where much work still remains to be done.

CAs will most likely become a game-changing technology that can offer
many benefits, provided that the issues just mentioned are carefully
considered. This technology may radically change the manner in which we
interact with machines, and will hopefully be developed in
an inclusive manner allowing, for example, disadvantaged groups the
same access to online and digital services as everyone else.
Because of the transformative nature of this technology,
making specific long-term predictions is very difficult. Thus,
we end this chapter by simply quoting Alan Turing\cite{Turing1950}: \lq\lq\textit{We 
can only see a short distance ahead, but we can see plenty there that
needs to be done.}\rq\rq

\bibliographystyle{ws-rv-van} 
\bibliography{arxiv_version}


\end{document}